\documentclass{article}
\usepackage{amsmath}
\usepackage{algorithm}
\usepackage{algpseudocode}
\usepackage{spconf,amsmath,graphicx}
\usepackage{float}
\usepackage{placeins}
\usepackage{caption}
\usepackage{makecell}
\usepackage{lipsum}
\usepackage[fontsize=9pt]{fontsize}

\usepackage{color}
\newcommand{\magenta}[1]{\textcolor{magenta}{#1}}

\usepackage{balance}
\usepackage{enumitem}
\usepackage{amssymb}
\usepackage{multirow}
\usepackage{amsfonts}
\usepackage{bbding}
\usepackage[numbers,sort&compress]{natbib}

\usepackage{url}

\let\OLDthebibliography\thebibliography
\renewcommand\thebibliography[1]{
  \OLDthebibliography{#1}
  \setlength{\parskip}{4pt}
  \setlength{\itemsep}{0pt plus 0.3ex}
}

\title{Multimodal Transformer Distillation for Audio-Visual Synchronization}
\name{Xuanjun Chen$^{12}$, Haibin Wu$^{1}$, Chung-Che Wang$^{2}$, Hung-yi Lee$^{1\dagger}$, Jyh-Shing Roger Jang$^{2\dagger}$\thanks{${\dagger}$ Equal correspondence. This work was supported by the National Science and Technology Council, Taiwan (Grant no. NSTC 112-2634-F-002-005). We also thank to National Center for High-performance Computing (NCHC) of National Applied Research Laboratories (NARLabs) in Taiwan for providing computational and storage resources. Code has been
made available at: \magenta{https://github.com/xjchenGit/MTDVocaLiST}.}}

\address{
$^1$Graduate Institute of Communication Engineering, National Taiwan University \\
$^2$Department of Computer Science and Information Engineering, National Taiwan University\\
\{d12942018, f07921092, hungyilee\}@ntu.edu.tw, geniusturtle6174@gmail.com, jang@mirlab.org}

\begin{document}
%
\maketitle

\begin{abstract}


Audio-visual synchronization aims to determine whether the mouth movements and speech in the video are synchronized.  VocaLiST reaches state-of-the-art performance by incorporating multimodal Transformers to model audio-visual interact information. However, it requires high computing resources, making it impractical for real-world applications. 
This paper proposed an MTDVocaLiST model, which is trained by our proposed multimodal Transformer distillation (MTD) loss. MTD loss enables MTDVocaLiST model to deeply mimic the cross-attention distribution and value-relation in the Transformer of VocaLiST. 
Additionally, we harness uncertainty weighting to fully exploit the interaction information across all layers.
Our proposed method is effective in two aspects:
From the distillation method perspective, MTD loss outperforms other strong distillation baselines. From the distilled model's performance perspective: 1) MTDVocaLiST outperforms similar-size SOTA models, SyncNet, and Perfect Match models by 15.65\% and 3.35\%; 2) MTDVocaLiST reduces the model size of VocaLiST by 83.52\%, yet still maintaining similar performance.

\end{abstract}


\section{Introduction}
\label{sec:intro}

The audio-visual synchronization task is to determine whether the mouth movements and speech in the video are synchronized.
An out-off-sync video may cause errors in many tasks, such as audio-visual user authentication \cite{aides2016text}, dubbing \cite{prajwal2020lip}, lip reading \cite{chung2019perfect}, active speaker detection \cite{tao2021someone, 10022646}, and audio-visual source separation \cite{afouras2020self, owens2018audio, zhu2022visually, pan2022selective}. 
An audio-visual synchronization model often acts as an indispensable front-end model for these downstream tasks. 
Various downstream tasks often run on mobile devices and require small model sizes and fast inference speed. Smaller model sizes and faster inference speed are required to ensure user experience, such as correcting the synchronization error of user-generated videos on mobile phones or performing audio-visual user authentication on finance mobile applications \cite{aides2016text}.
To work with these applications, a lightweight audio-visual synchronization model is worth exploring.

A typical framework for audio-visual synchronization tasks is estimating the similarity between audio and visual segments.
SyncNet \cite{chung2016out} introduced a two-stream architecture to estimate the cross-modal feature similarities, which is trained to maximize the similarities between features of the in-sync audio-visual segments and minimize the similarities between features of the out-of-sync audio-visual segments. Perfect Match (PM) \cite{chung2019perfect, chung2020perfect} optimizes the relative similarities between multiple audio features and one visual feature with a multi-way matching objective function. Audio-Visual Synchronisation with Transformers (AVST) \cite{chen2021audio} and VocaLiST \cite{kadandale2022vocalist}, the current state-of-the-art (SOTA) models, which incorporate Transformers \cite{vaswani2017attention} to learn the multi-modal interaction and classify directly if a given audio-visual pair is synchronized or not, resulting in an excellent performance. 
However, both AVST and VocaLiST require large memory and high computing costs, making these models unsuitable for edge-device computation.
 
In this paper, we propose to distill a small-size version of VocaLiST, namely MTDVocaLiST, which is distilled by mimicking the multimodal Transformer behavior of VocaLiST. 
Next, we propose to employ uncertainty weighting, which allows us to assess the varying significance of Transformer behavior across different layers, resulting in an enhanced MTDVocaLiST model. 
To our knowledge, this is the first attempt to distill a model by mimicking multimodal Transformer behavior for the audio-visual task.
Our model outperforms similar-size state-of-the-art models, SyncNet and PM models by 15.65\% and 3.35\%. 
MTDVocaLiST significantly reduces VocaLiST's size by 83.52\% while still maintaining competitive performance comparable to that of VocaLiST.

\begin{figure*}
\centering
\includegraphics[width=17.8cm]{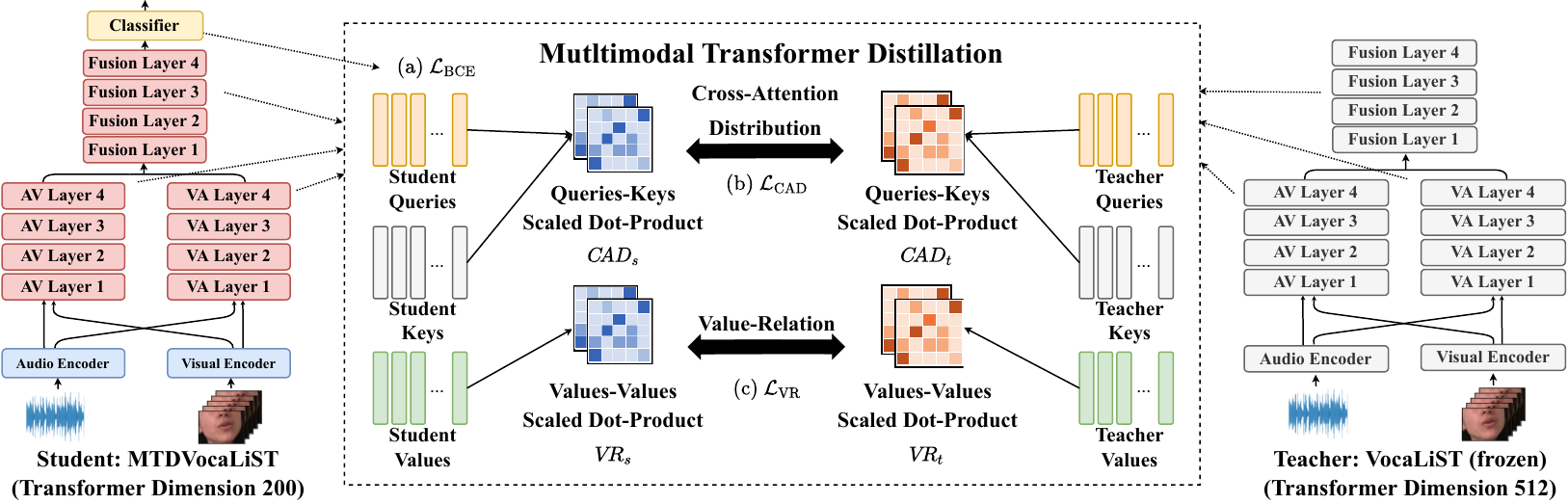}
\vspace{-2em}
\caption{The proposed MTDVocaLiST model. (a) binary cross entropy loss. (b) cross-attention distribution distillation loss. (c) value-relation distillation loss. }

\label{fig:MTDVocaLiST}
\end{figure*}

\section{Background}
\label{sec:methods}

\subsection{VocaLiST}
\label{sec: vocalist}

VocaLiST \cite{kadandale2022vocalist} is a SOTA audio-visual synchronization model. 
The input of VocaLiST is a sequence of visuals and its corresponding audio features. 
The output is about to classify whether a given audio-visual pair is in sync. VocaLiST consists of an audio-visual front-end and a synchronization back-end. 
The audio-visual front end extracts audio and visual features. The synchronization back-end comprises three cross-modal Transformer encoders, namely audio-to-visual (AV) Transformer,  visual-to-audio (VA) Transformer, and Fusion Transformer. Each Transformer block has 4 layers. The core part of the cross-modal Transformer is the cross-attention layer, whose input has queries, keys, and values.
In VocaLiST, the AV Transformer uses audio for queries and visual data for keys and values, while the VA Transformer does the opposite. The Fusion Transformer merges these, taking AV output for queries and VA output for keys and values. Its output undergoes max-pooling over time and is activated with a tanh function. A fully connected layer then classifies if voice and lip motion are synchronized, using binary cross-entropy loss for optimization.

\subsection{Knowledge Distillation}
\label{sec: knowledge}

Naïve knowledge distillation (KD) \cite{hinton2015distilling} will let the student learn temperature-controlled soft targets from the teacher's output.
However, the output representations of the intermediate layers have also been shown to be able to guide the training of student models \cite{romero2014fitnets}. 
Relational knowledge distillation (RKD) \cite{park2019relational} also finds rich relational knowledge between different data samples. RKD transfers instance relations modeled by distance and angle relational potential functions. Besides, a series of works \cite{zagoruyko2016paying, tung2019similarity, peng2019correlation, ahn2019variational, passalis2018learning, heo2019knowledge, kim2018paraphrasing, yim2017gift, huang2017like, tian2019crd, chen2021wasserstein,liao22_spsc} also aims to distill different knowledge sources based on the teacher-student framework. 
However, none of the above distillation methods have taken into account the characteristics of the Transformer encoder. 
In natural language processing, MiniLM \cite{wang2020minilm} attempts to mimic the self-attention of the last Transformer encoder layer from a large self-supervised learning model, which achieves impressive results. 

\section{MTDVocaLiST}
\label{sec:mtdvocalist}

Inspired by MiniLM, we propose a multimodal Transformer distillation VocaLiST (MTDVocaLiST) model, as shown in Fig. \ref{fig:MTDVocaLiST}. The basic idea is to encourage the student model MTDVocaLiST to learn multi-modal interaction behavior from the teacher model VocaLiST in cross-modal Transformers.
Prior to the distillation process, both MTDVocaLiST and VocaLiST will extract their respective Transformer behaviors, specifically Cross-Attention Distribution (CAD) and Value-Relation (VR). 
The loss function for training MTDVocaLiST incorporates the binary cross-entropy loss $\mathcal{L}_{BCE}$ and the Transformer behavior mimic losses (i.e., Equation (\ref{eq:TransBehaviorloss})).
The Transformer behavior mimic losses, denoted as $\mathcal{L}_{CAD}$ and $\mathcal{L}_{VR}$, are designed to facilitate the student's learning from the teacher. 
During the distillation, VocaLiST is frozen, and we only train MTDVocaLiST. 
The teacher uses the public pre-trained model of VocaLiST\footnote{https://github.com/vskadandale/vocalist} for initialization. 
MTDVocaLiST has a similar architecture to VocaLiST, but its embedding dimension is reduced from 512 to 200. 
MTDVocaLiST and VocaLiST have the same number of layers.
VocaLiST has 80.1 million parameters, while MDTVocaLiST has only 13.2 million parameters, reducing the size of the teacher model by 83.52\%.

\noindent\textbf{Multimodal Transformer Behaviors.}
In our framework, Transformer behavior encompasses CAD and VR. 
CAD measures the attention relationship between one modal (e.g., audio) with another modal (e.g., visual) in a multimodal task. 
VR conducts scaled dot product operations among features within the same modality to capture their correlations and dependencies. This enhances the model's understanding of relationships between different parts within the same modality. 
The formula for CAD is the same as original attention \cite{vaswani2017attention}, but we use scaled dot-product value to model the VR instead of just the attention values, which has proven to be more effective \cite{wang2020minilm}.  
The CAD and VR are formulated as follows:
\begin{equation}
\label{eq:TransBehavior}
\begin{aligned}
C\!A\!D = \text{softmax}\left(\frac{Q K^T}{\sqrt{d_q}}\right),  \quad
V\!R = \text{softmax}\left(\frac{V V^T}{\sqrt{d_v}}\right)
\end{aligned}
\end{equation}
where queries $Q$, keys $K$, and values $V\!$ depend on the type of Transformer block. 
$d_q$ and $d_v$ represent the number of dimension of $Q$ and $V\!$.
We formulate the loss function of learning the Transformer behavior as follows:
\begin{equation}
\label{eq:TransBehaviorloss}
\begin{aligned}
\mathcal{L}_{C\!A\!D} = D_{KL} ({C\!A\!D}_{s} || {C\!A\!D}_{t}), \quad
\mathcal{L}_{V\!R} = D_{KL} ({V\!R}_{s} || {V\!R}_{t})
\end{aligned}
\end{equation}
where we denote the Transformer behavior $C\!A\!D_{t}$, $V\!R_{t}$ produced by the teacher's layer via Equation (\ref{eq:TransBehavior}) and its corresponding student's Transformer behavior are ${C\!A\!D}_{s}$, $V\!R_{s}$. $D_{KL}$ is the Kullback–Leibler divergence. 
The Fusion, AV and VA Transformer blocks of the VocaLiST are key blocks for modeling different audio-visual interactions. 
Since each Transformer layer contains unique information, we've developed naïve multimodal Transformer distillation (NMTD) and multimodal Transformer distillation (MTD) to effectively harness these diverse Transformer layers.

\noindent\textbf{Naïve Multimodal Transformer Distillation (NMTD).}
To enable MTDVocaLiST to comprehensively learn multimodal interaction information at various levels, we aggregate the CAD and VR losses from each layer using weighted sums. 
Combined with Equation (\ref{eq:TransBehaviorloss}), our NMTD loss can be formulated as follows:
\begin{equation}
\label{eq:NaiveMTD}
\begin{aligned}
\mathcal{L}_{NMTD} = w_0 \cdot \mathcal{L}_{BCE} + \sum^{L}_l w_{l1} \cdot \mathcal{L}_{{C\!A\!D}_l} +  w_{l2} \cdot \mathcal{L}_{{V\!R}_l},
\end{aligned}
\end{equation}
where $L$ represents a candidate set of layers. $w_{l1}$ and $w_{l2}$ refer to the weights of CAD and VR loss for the $l^{th}$ layer. $w_0$ is a weight used to control $\mathcal{L}_{BCE}$. 
There are three common methods for choosing loss weights for Equation (\ref{eq:NaiveMTD}): uniform, manual tuning, and automatic weighting (AW).
Uniform weights assign weight one to all terms. However, this approach ignores it there may be some conflict relations between losses.
Manual tuning relies on human heuristics to find optimal loss weights but involves trial-and-error. 
AW utilizes learnable weights for automatic weighting, but it may lead to weight vanishing and requires gradient clipping.

\noindent\textbf{Mutltimodal Transformer Distillation (MTD).}
To address the weighting issues, we further introduced the uncertainty weighting (UW) method to improve the student model.
MTDVocaLiST's primary objective was to mimic three distinct types of multimodal Transformer layers, resulting in a total of 12 layers. 
However, the performance contribution of each layer or type of Transformer to MTDVocaLiST is uncertain. 
Previous research has shown that employing Bayesian modeling of uncertainty is effective in balancing loss contributions in multi-task learning \cite{kendall2017multi, liebel2018auxiliary}.
Therefore, we can regard the complete multimodal Transformer distillation framework as encompassing multiple distillation tasks alongside a supervised learning task. 
After utilizing uncertainty weighting, the overall MTD can be formulated as follows:
\begin{equation}
\label{eq:UW_MTD}
\begin{aligned}
\mathcal{L}_{MTD} = {\sum^{T}_{\tau}} \frac{1}{{2} \cdot {w^2_\tau}} \cdot {\mathcal{L}_{\tau}} + ln(1+w^2_\tau),
\end{aligned}
\end{equation}
where $T$ represents a task set, and $\mathcal{L}_\tau$ denotes the loss for the $\tau\text{-}th$ task. 
Within task sets $T$, there is a scope for $\mathcal{L}_{BCE}$, along with losses pertaining to various layers of $\mathcal{L}_{CAD}$ and $\mathcal{L}_{VR}$. 
The loss of each task is associated with its specific learnable parameter $w_\tau$. 
The term $ln(1 + w^2_\tau)$ serves the purpose of enforcing positive regularization values.
The overarching objective can be viewed as learning the relative weights assigned to different tasks. When $w_\tau$ assumes large-scale values, it diminishes the contribution of $\mathcal{L}_{\tau}$, while small-scale values amplify its contribution.

\section{Experiment}
\label{sec:exp}


\subsection{Experimental setup}

\textbf{Model and dataset.}
The input of MTDVocaLiST is an audio-visual pair, which corresponds to a video sequence of 5 frames (0.2 seconds) sampled at 25 frames per second. 
We also train the model with only $\mathcal{L}_{BCE}$ loss using the MTDVocaLiST architecture as the baseline. 
All models are trained on the LRS2 dataset \cite{afouras2018deep}, which contains 96,318 utterances in the pretraining set, 1,082 utterances in the validation set, and 1,243 utterances in the testing set. The maximum number of characters in one utterance is 100.

\noindent\textbf{Training and validation.}
During the process, positive and negative samples are equally sampled on the fly. Positive samples mean synchronized audio-visual pairs.
Negative samples are obtained by introducing random temporal misalignment at offsets within a time scale of ±15 visual frames (1.2s) between the in-sync audio-visual pairs. 
The model with the highest F1 score in validation is saved for evaluation.
We train for a total of 80 epochs, which is significantly less than the 600 epochs required for VocaLiST training.

\noindent\textbf{Evaluation protocol.} 
The evaluation protocol is similar to the cross-modal retrieval task, which follows previous work \cite{chung2016out, chen2021audio, chung2019perfect, chung2020perfect,kadandale2022vocalist}.
Given a 5-frame visual segment and a candidate audio set containing 31 audio segments, the model will predict the corresponding score for each pair of audio-visual segments. 
The candidate set consists of the target audio segment and 15 left and right neighbors. 
The target is to find the index of an audio segment most similar to the given 5-frame visual segment. 
The found index is considered correct if the offset between it and the ground-truth is within the $\pm1$ frame since human beings distinguish no difference within the $\pm1$ frame. 
Since a short input frame might not be enough to determine the correct offset \cite{chung2016out}. 
Thus, we also conduct experiments using input frame lengths over 5 frames of video (i.e., 5, 7, 9, 11, 13, and 15), averaging the prediction scores over multiple video samples (with a 1-frame temporal stride). 
In the ablation study, we limit the evaluation to an input frame length of 5 frames due to page space constraints.


\begin{table}[t]
\renewcommand{\tabcolsep}{3.5pt}
\renewcommand\arraystretch{1.2}
\small
\vspace*{-0.4cm}
\caption{Accuracy of different distillation methods in evaluation.}
\centerline{
\begin{tabular}{c|c|c|c|c|c|c}
\hline
\multirow{3}{*}{\shortstack[c]{Distillation\\ method}} & \multicolumn{6}{c}{Input frame length (seconds)} \\
\cline{2-7}
           &  \multirow{2}{*}{\shortstack[c]{5 \\ (0.2s)}}  &  \multirow{2}{*}{\shortstack[c]{7 \\ (0.28s)}}    &    \multirow{2}{*}{\shortstack[c]{9 \\ (0.36s)}}    &  \multirow{2}{*}{\shortstack[c]{11 \\ (0.44s)}}      & \multirow{2}{*}{\shortstack[c]{13 \\ (0.52s)}}      & \multirow{2}{*}{\shortstack[c]{15 \\ (0.6s)}} \\
           &       &   &   &   &  &  \\

\hline
\hline
$\mathcal{L}_{BCE}$ &  71.36 &   81.44   &   88.84   &   93.41   &   96.19   &  97.69  \\
 KD  &  80.87 &   88.62   &   93.48   &   96.32   &   97.90   &  98.82  \\
RKD  &  86.06 &   92.42   &   95.95   &   97.80   &   98.75   &  99.29  \\
MiniLM$^*$  &  85.60 &   92.03   &   95.91   &   97.72   &   98.72   &  99.25  \\

FitNets &  90.81 &   95.48   &   97.77   &   98.81   &   99.42   &  99.66  \\


\hline
MTD    &   \textbf{91.45}  &  \textbf{95.75}  &   \textbf{97.99}   &  \textbf{98.95}  &  \textbf{99.46}  &  \textbf{99.68}  \\  
\hline

\end{tabular}}

\label{tab:acc_of_diff_kd}
\end{table}

\begin{figure}
\centering
\includegraphics[width=8.5cm]{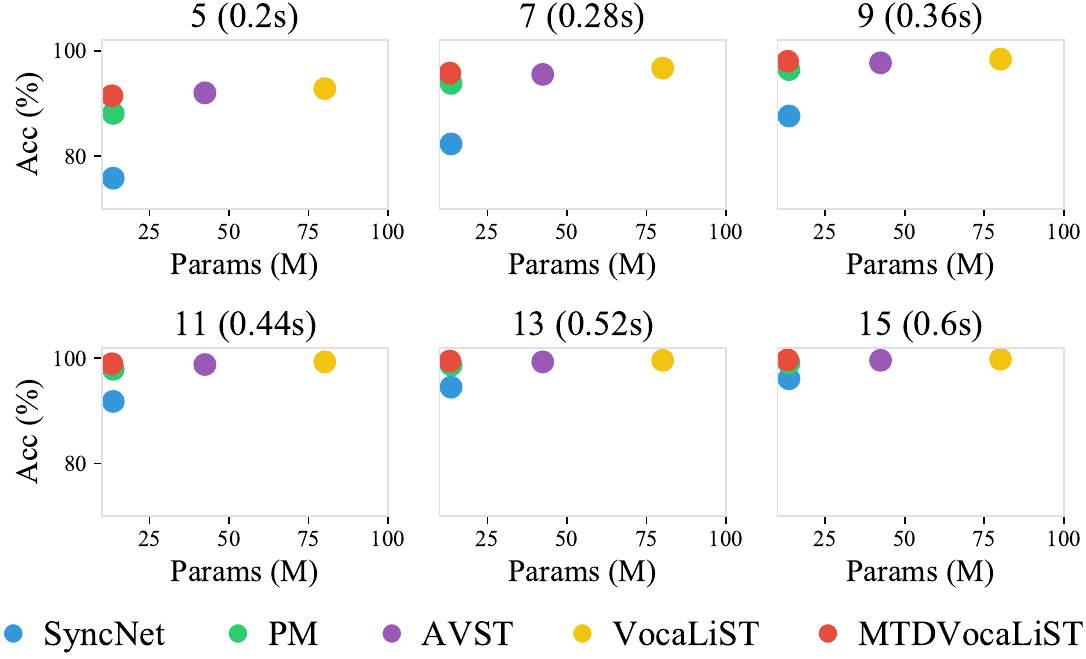}
\vspace*{-0.2cm}
\caption{Comparison of model size and accuracy.}

\label{fig:comparison_with_sota}

\end{figure}


\subsection{Main Results}
\label{sec:MainResults}

\noindent\textbf{Comparison with Different Distillation Methods.} 
We compare MTD loss with several other knowledge distillation methods, including knowledge distillation (KD) \cite{hinton2015distilling}, relational knowledge distillation (RKD) \cite{park2019relational}, MiniLM*, and FitNets.
As training MiniLM directly does not converge, we use the loss of MiniLM combined with $\mathcal{L}_{BCE}$ for training, denoted as MiniLM*. FitNets uses the last layer representation as hints and guide layers \cite{romero2014fitnets}. We replace the MTD loss of MTDVocaLiST with different losses for comparison. Table \ref{tab:acc_of_diff_kd} demonstrates that MTD outperforms other distillation methods for most input frame lengths. 
When considering an input frame length of 5, training solely with $\mathcal{L}_{BCE}$ results in the lowest accuracy at 71.36\%. In contrast, MTD loss significantly improves accuracy, outperforming KD by 10.58\%, RKD by 5.39\%, MiniLM* by 5.85\%, and FitNets by 0.64\%.
Similar trends are observed across different input frame lengths. 

\noindent\textbf{Comparison with SOTA models.}
We also compare MTDVocaLiST with previous SOTA models, 
including SyncNet, PM, AVST, and VocaLiST. 
Fig. \ref{fig:comparison_with_sota} presents a graphical depiction of the relationship between model size and accuracy across various input frame lengths.  
We present the following observations:
1) In equivalent model size configurations, MTDVocaLiST outperforms SyncNet and PM significantly across all frame settings. Notably, MTDVocaLiST surpasses SyncNet and PM by 15.65\% and 3.35\%, respectively, when utilizing a 5-frame input length.
2) Remarkably, with only 31.13\% of the model size of AVST, MTDVocaLiST achieves superior performance in most input frame length scenarios.
3) Furthermore, MTDVocaLiST utilizes only 16.48\% of VocaLiST's model size while maintaining comparable accuracy. In summary, across all input frame lengths, MTDVocaLiST demonstrates competitive performance against VocaLiST and surpasses AVST, SyncNet, and PM.

\begin{table}[t]
\renewcommand\arraystretch{1.2}
\small
\caption{Ablation study of NMTD loss.}
\centerline{
\begin{tabular}{c|c|c}
\hline
\multirow{1}{*}{\shortstack[c]{Loss}} & \multirow{1}{*}{\shortstack[c]{Val F1 (\%)}} &  \multirow{1}{*}{\shortstack[c]{Eval Acc (\%)}} \\
\hline
\hline
$\mathcal{L}_{BCE}$ & 87.91  &  71.36  \\
NMTD w/o $\mathcal{L}_{VR}$ & 91.78  & 83.55  \\
NMTD w/o $\mathcal{L}_{CAD}$ & 91.97  & 83.53  \\
NMTD &  \textbf{92.81}  & \textbf{85.60}  \\
\hline
\end{tabular}}
\label{tab:mtd_loss}
\end{table}

\begin{figure}[t]
\centering

\vspace{-0.5em}
\includegraphics[width=8.5cm]
{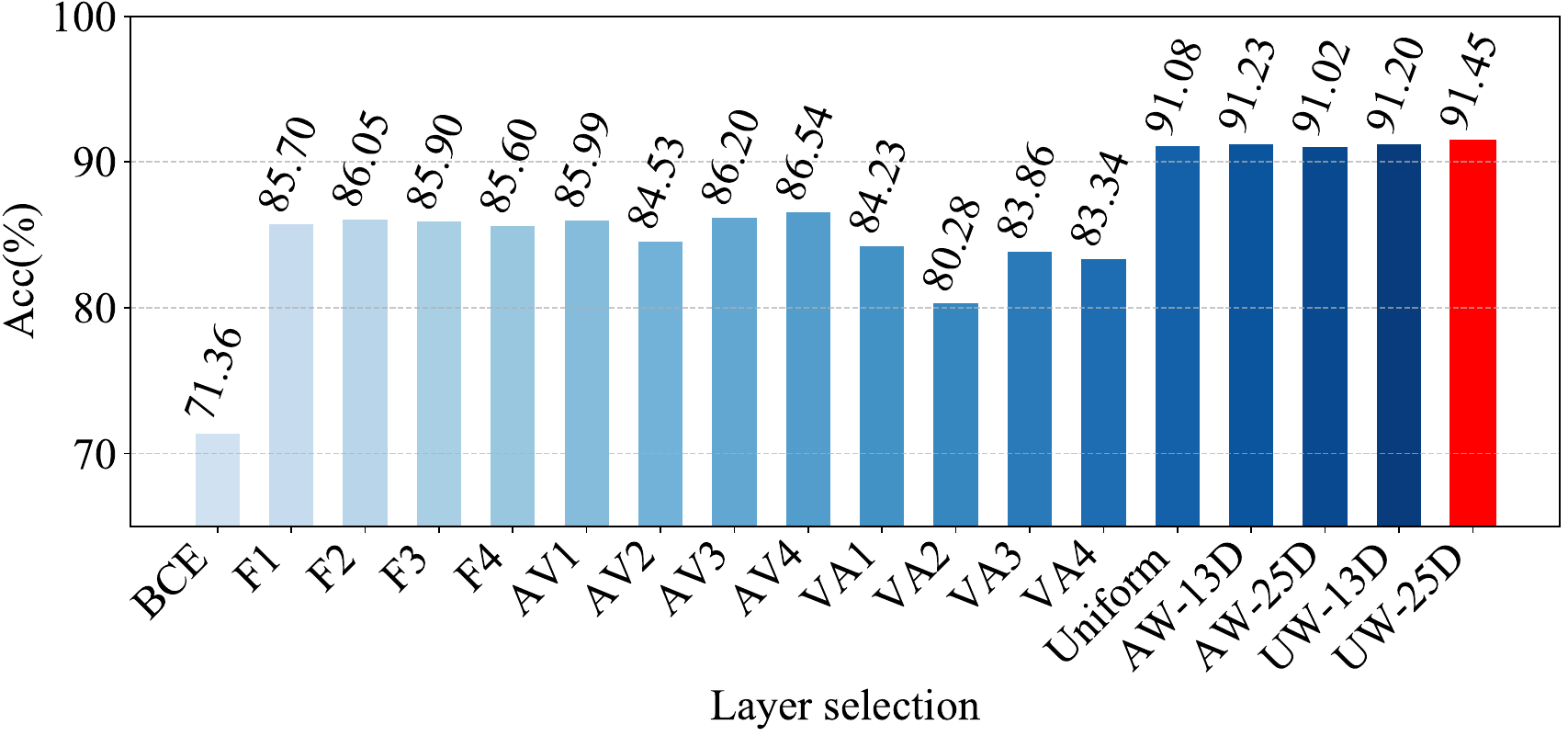}
\vspace*{-0.3cm}
\caption{\textit{}Accuracy of different layer selection strategies.}
\label{fig:abl_layer_selection}

\end{figure}









\subsection{Ablation study and analysis}
\label{subsec:ablation}

\noindent\textbf{Indispensability of CAD and VR.}
To better understand the significance of $\mathcal{L}_{CAD}$ and $\mathcal{L}_{VR}$, we conducted an ablation study for NMTD loss on the last layer of the fusion Transformer block (i.e., fusion layer 4). We only use uniform weighting.
In Table \ref{tab:mtd_loss}, the results reveal that both cross-attention distribution and value-relation contribute significantly to NMTD loss, which outperforms $\mathcal{L}_{BCE}$.

\noindent\textbf{Layer selection.}
In Fig. \ref{fig:abl_layer_selection}, we compare different training strategies, including BCE, single fusion Transformer layers (F1 $\sim$ F4), single AV Transformer layers (AV1 $\sim$ AV4), single VA Transformer layers (VA1 $\sim$ VA4), and layer weighting methods (Uniform, AW-13D, AW-25D, UW-13D, UW-25D). 
Most implementations are based on NMTD loss except UW which is based on MTD loss. 
For single-layer distillation and BCE training using Equation (\ref{eq:NaiveMTD}), we can manually set the weight of specific terms to 1 while setting the weights of the other terms to 0. 
The 13D and 25D suffixes denote the use of 13-dimensional and 25-dimensional learnable parameters, respectively. 
In the former case, $\mathcal{L}_{CAD}$ and $\mathcal{L}_{VR}$ losses within each layer share the same parameter, while in the latter, each $\mathcal{L}_{CAD}$ and $\mathcal{L}_{VR}$ losses in different Transformer layers has its own learnable parameter. 
Employing more learnable parameters will result in better performance.

Fig. \ref{fig:abl_layer_selection} yields the following insights:
1) Distilling any Transformer layer significantly improves performance compared to using only BCE loss for supervised learning.
2) Distilling VA Transformer layers consistently performs worse than distilling fusion or AV Transformer layers, suggesting that VA layers contribute minimally to the student's final performance.
3) Models Uniform, AW-13D, AW-25D, UW-13D, and UW-25D consistently outperform single-layer distillation and BCE training. 
The greater the number of layers being distilled, the better the performance.
4) Regarding layer weighting, AW-13D outperforms Uniform by 0.15\%, while AW-25D is 0.06\% lower than Uniform. In contrast, UW-13D and UW-25D exhibit more stable performance, with UW-25D achieving 91.45\%. Notably, our UW method does not require manual tuning. 
Therefore, we employed the UW-25D layer selection strategy for the best MTDVocaLiST in section \ref{sec:MainResults}.

\noindent\textbf{Transformer Behavior and Transformer Representation.} 
Both FitNets and MTD losses aim to distill Transformer layers, with MTD loss consistently delivering superior performance in Table \ref{tab:acc_of_diff_kd}. 
To delve deeper into this phenomenon, we conduct further analysis. 
A typical Transformer layer mainly consists of a multi-head attention layer and a multi-layer perceptron layer. 
The output of the multi-layer perceptron layer is also known as Transformer representations. 
There are some differences between MTD loss and FitNets. 
MTD loss aims to mimic the multi-head attention mechanism, which we refer to as Transformer behavior. 
FitNets loss aims to learn Transformer representations from the last distilled layers. 

\begin{figure}[t]
\centering

\includegraphics[width=8.5cm]{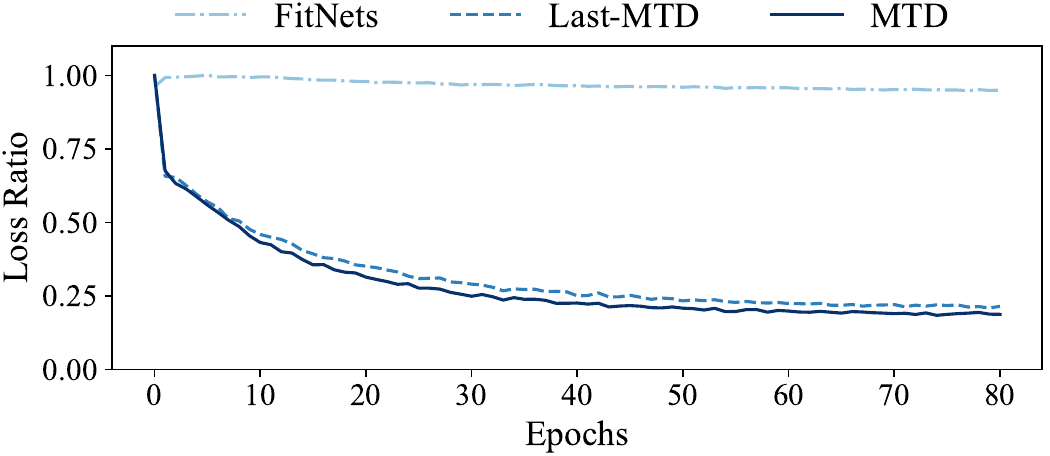}
\vspace*{-0.3cm}
\caption{Comparison of Transformer representation and its behavior.}


\label{fig:rep_and_qkv}
\vspace*{-0.5cm}
\end{figure}

We observe the relationship between losses through Fig. \ref{fig:rep_and_qkv}, where the x-axis corresponds to epochs and the y-axis indicates the loss ratio. 
The loss ratio means that losses are normalized by their respective maximum values across all epochs. 
In Fig. \ref{fig:rep_and_qkv}, our model was solely trained using the MTD loss. To make a fair comparison with FitNets, we also propose a Last-MTD loss, which solely utilized $\mathcal{L}_{BCE}$ loss and the last layer Transformer behavior loss. 
Fig. \ref{fig:rep_and_qkv} represents the loss computation during the validation, where we concurrently calculated the losses for MTD, Last-MTD, and FitNets. 
We monitored how these losses evolved as the number of epochs increased. 
Notably, when we focused on optimizing solely the MTD loss, the Last-MTD loss exhibited a concurrent decrease along with the MTD loss. In contrast, the FitNets loss did not demonstrate a significant decrease. 
This observation suggests that what is distilled from the representation layers and what is distilled from imitating the Transformer behavior are distinct aspects of the Transformer. 

\section{Conclusion}
\label{sec:con}

This work introduces MTDVocaLiST, a model that deeply learns to mimic the cross-attention distribution and value-relation of VocaLiST. 
Enhanced by uncertainty weighting, our MTD loss utilizes the relative importance of Transformer behavior across various layers, leading to an improved model.
Experimental results show that MTD loss outperforms other strong distillation baselines, and MTDVocaLiST maintains competitive performance while reducing the teacher model's size by 83.52\%. Notably, our model outperforms SyncNet and PM models of similar sizes by 15.65\% and 3.35\%, respectively.

\ninept
\section{References}
\bibliographystyle{IEEEbib}

\bibliography{refs}
\vfill\pagebreak

\end{document}